\documentclass[10pt,twocolumn,letterpaper]{article}

\usepackage[pagenumbers]{cvpr} 

\usepackage{graphicx}
\usepackage{amsmath}
\usepackage{amssymb}
\usepackage{booktabs}

\usepackage{algorithm}
\usepackage{algorithmic}

\usepackage[normalem]{ulem}
\useunder{\uline}{\ul}{}
\usepackage{multirow}
\usepackage{booktabs}
\usepackage[pagebackref,breaklinks,colorlinks]{hyperref}

\usepackage[capitalize]{cleveref}
\crefname{section}{Sec.}{Secs.}
\Crefname{section}{Section}{Sections}
\Crefname{table}{Table}{Tables}
\crefname{table}{Tab.}{Tabs.}

\def\cvprPaperID{*****} 
\def\confName{CVPR}
\def\confYear{2023}

\begin{document}

\title{Explaining Cross-Domain Recognition with Interpretable Deep Classifier}

\author{
  Yiheng~Zhang, Ting~Yao, Zhaofan~Qiu, and~Tao~Mei \\
  JD Explore Academy, Beijing, China \\
{\tt\small \{yihengzhang.chn, tingyao.ustc, zhaofanqiu\}@gmail.com, tmei@jd.com}
}
\maketitle

\begin{abstract}
The recent advances in deep learning predominantly construct models in their internal representations, and it is opaque to explain the rationale behind and decisions to human users. Such explainability is especially essential for domain adaptation, whose challenges require developing more adaptive models across different domains. In this paper, we ask the question: how much each sample in source domain contributes to the network's prediction on the samples from target domain. To address this, we devise a novel Interpretable Deep Classifier (IDC) that learns the nearest source samples of a target sample as evidence upon which the classifier makes the decision. Technically, IDC maintains a differentiable memory bank for each category and the memory slot derives a form of key-value pair. The key records the features of discriminative source samples and the value stores the corresponding properties, e.g., representative scores of the features for describing the category. IDC computes the loss between the output of IDC and the labels of source samples to back-propagate to adjust the representative scores and update the memory banks. Extensive experiments on Office-Home and VisDA-2017 datasets demonstrate that our IDC leads to a more explainable model with almost no accuracy degradation and effectively calibrates classification for optimum reject options. More remarkably, when taking IDC as a prior interpreter, capitalizing on 0.1\% source training data selected by IDC still yields superior results than that uses full training set on VisDA-2017 for unsupervised domain adaptation.
\end{abstract}

\section{Introduction}
One important factor credited for the remarkable developments in computer vision today is the emergence of deep neural networks. Despite having encouraging performances, researchers start to see some downsides of deep learning methods, e.g., needing large volumes of big data, computational power and engineering efforts of human experts. More importantly, these approaches become increasingly opaque to the end users. This lack is very brittle against real-world deployment of more intelligent and autonomous systems. As such, explainable artificial intelligence (XAI) is gaining intensive traction recently and refers to the techniques to build interpretable systems whose decisions can be understood. Existing methods predominantly explain the deep model's decisions in the context of in-domain understanding through the use of saliency or attention maps~\cite{Petsiuk2018rise,selvaraju2020grad,simonyan2013deep,springenberg2015striving,TMM_EXP_Wang}. In this paper, we expand the horizons of XAI to a more challenging scenario of cross-domain recognition, particularly in explaining unsupervised domain adaptation, which transfers the knowledge learnt from the source domain with labeled examples to the target domain with only unlabeled data.

\begin{figure}[!tb]
    \centering\includegraphics[width=0.95\linewidth]{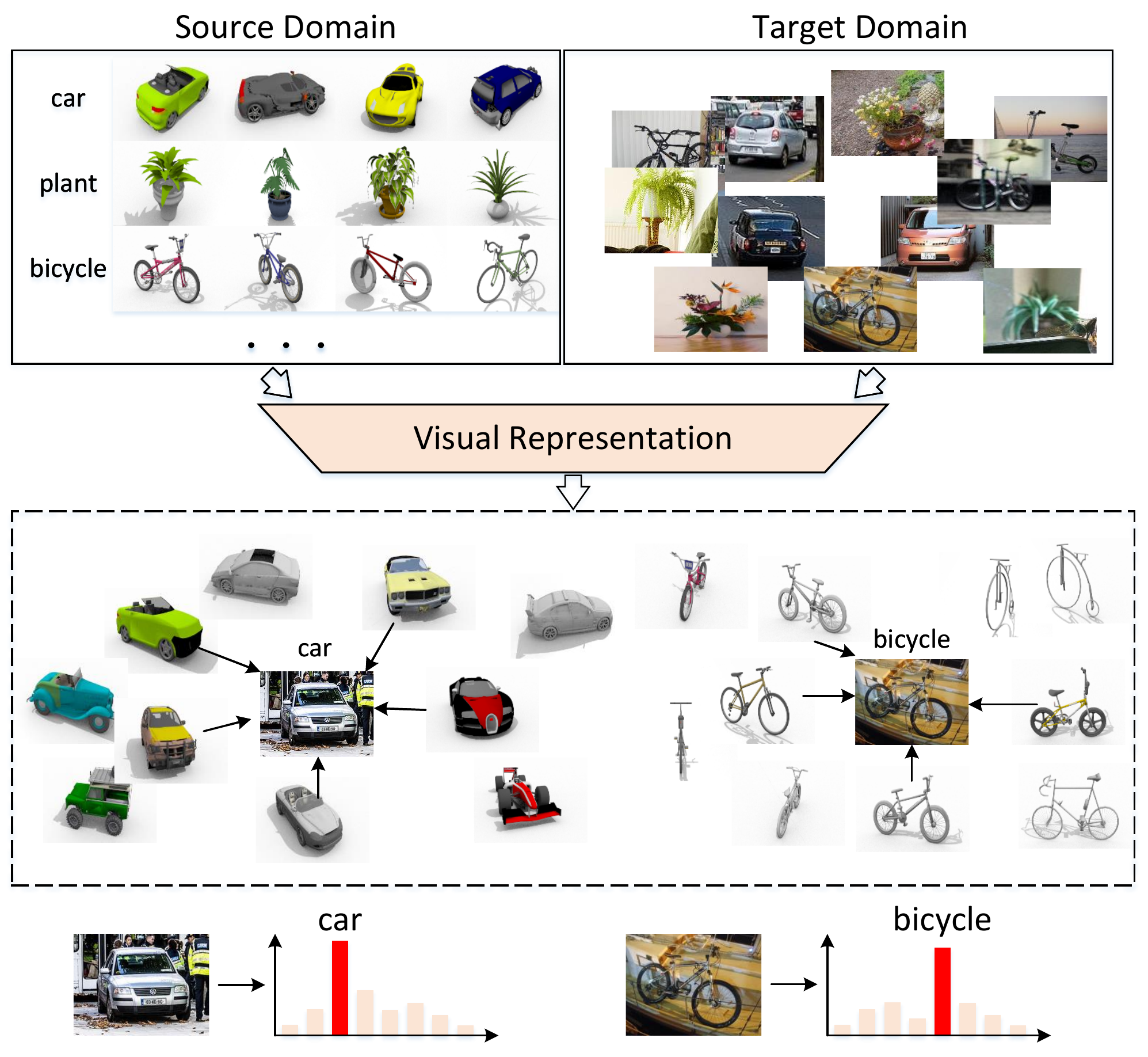}
    \vspace{-0.1in}
    \caption{\small The rationale behind the cross-domain classification of $k$-NN, which is based on distance measure.}
    \label{fig1:1}
    \vspace{-0.2in}
\end{figure}

\begin{figure*}[t]
    \centering\includegraphics[width=0.95\linewidth]{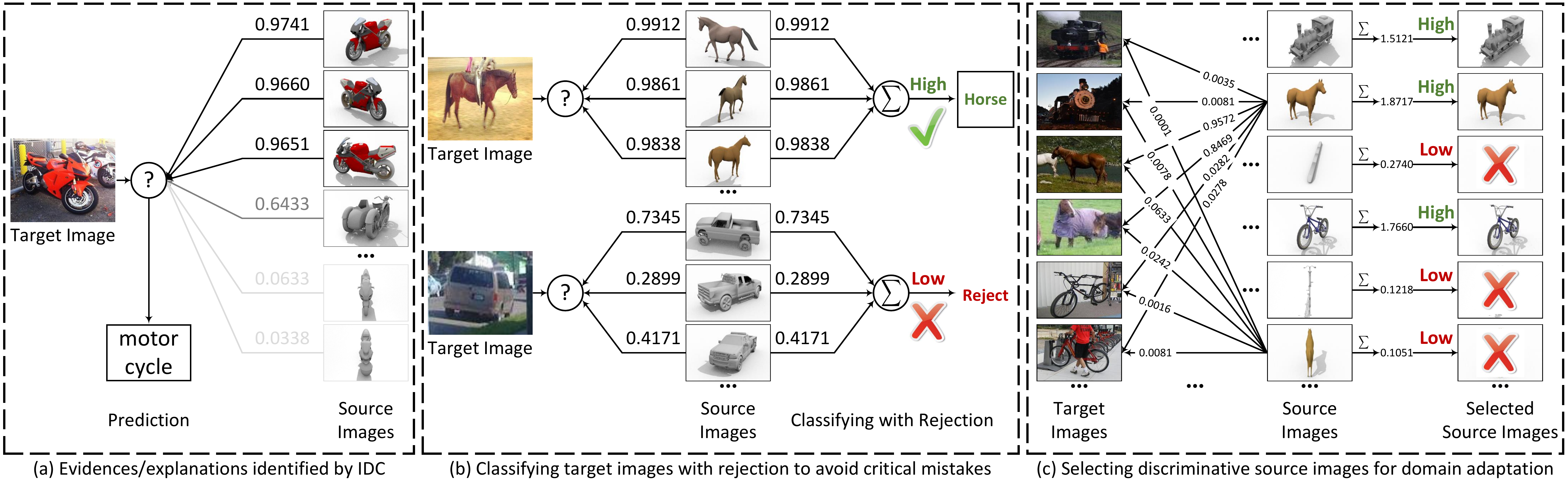}
    \vspace{-0.1in}
    \caption{\small (a) Identifying the highly similar keys from memory banks by IDC as the evidence or explanations for the prediction. (b) Refusing to make a decision when the confidence scores of the evidence by IDC are low. (c) Taking IDC as a prior interpreter to select discriminative source images with respect to the target domain.}
    \label{fig.tasks}
    \vspace{-0.2in}
\end{figure*}
    
In an effort to study the problem, we define the explanations here as the evidence of source examples dominantly upon which a deep classifier makes the prediction of target samples. In practice, we intend to borrow the high explainability of $k$-nearest neighbors ($k$-NN) to characterize the explanations. \Cref{fig1:1} conceptually depicts how the decision-making process will behave in an interpretable way. Given a set of labeled source images and unlabeled target images, the distances between each target image and source images are measured in feature space, and utilized as the explanations for indicating the contribution of each source image towards recognizing the target image. The classifier then categorizes the unlabeled target image based on the source images that are close to it. This simple yet effective way conveys the rationale behind the decision-making process. 
Thus, we propose to derive the spirit of distance learning in $k$-NN and devise an explainable deep classifier maintaining a high level of performance.

To materialize the idea, we present an Interpretable Deep Classifier (IDC) for unsupervised domain adaptation.
IDC designs a series of independent memory banks and each memory bank corresponds to one specific category. The memory slot is in the key-value form.
IDC places image features in the key and the value contains the corresponding properties, e.g., the learnable representative scores of the features for describing the category.
Given an input image, we extract the image representations as a query and IDC reads from memory bank to measure the similarity between each key and the query.
IDC takes the similarity as the weight and computes the weighted average representative score over highly similar keys in the memory bank as the probability score of the input image on this category.
The difference between the prediction of IDC and the ground-truth label is evaluated to back-propagate the gradients to update the representative scores in the memory banks. Note that because the images from target domain are unlabeled, the back-propagation only works on source images.
IDC then writes the query into the memory or replaces the least frequently accessed memory slot. We integrate our IDC into the widely-adopted adversarial learning framework, tailored for unsupervised domain adaptation.
In addition to IDC, the fully-connected layer guides representation learning on source images in a supervised manner and the domain discriminator encourages the learnt representations to be domain invariant from an adversarial perspective.

In summary, IDC decouples the decision-making process into evidence identification (identifying) and inference w.r.t the evidence (inferring). For making decisions, IDC reads from each memory bank to identify the highly similar keys, which are considered as evidence or explanations in our context (\Cref{fig.tasks}(a)), with respect to the test image from target domain. IDC then utilizes the similarity between the test image and these evidence to weighted average the representative scores of the evidence as the probability of the test image on each category. The test image is finally classified by assigning the category with the highest probability in the inference.
The evaluations of IDC are carried out on both Office-Home and VisDA-2017 datasets for unsupervised domain adaptation. IDC learns more effective explanations and also maintains a high accuracy. We further validate the explainability of IDC from two aspects.
One is to verify IDC on classifying the images in target domain with rejection (\Cref{fig.tasks}(b)), where IDC can choose not to make a decision to avoid critical mistakes.
The other is to take IDC as a prior interpreter to select the discriminative source images for domain adaptation (\Cref{fig.tasks}(c)). The results on VisDA show that learning with very few source data (e.g., 0.1\%) selected by IDC outperforms that trains on the full set, which is very impressive.

\section{Related Works}
\textbf{Unsupervised Domain adaptation} (UDA) is to alleviate domain shift between labeled source domain and unlabeled target domain. 
The advances have proceeded mainly along two dimensions: domain discrepancy minimization~\cite{long2015learning,long2017deep,saenko2010adapting,TMM_UDA_Yan_MMD,TMM_UDA_Lu_DM} and adversarial learning~\cite{cui2020gvb,ganin2016domain,long2018conditional,saito2018maximum,zhang2019bridging,TMM_UDA_Jing_ADV,TMM_UDA_Shermin_ADV}. 
The former is to learn transferable representations by minimizing domain discrepancy through Maximum Mean Discrepancy (MMD)~\cite{gretton2012kernel,tzeng2014deep}. 
The MMD-based frameworks are further improved in~\cite{long2015learning,long2017deep,long2016unsupervised} by residual transfer~\cite{long2015learning}, multi-kernel MMD~\cite{long2016unsupervised} and joint distributions~\cite{long2017deep}. 
The latter~\cite{ganin2015unsupervised,tzeng2017adversarial} follows the spirit behind Generative Adversarial Networks (GAN)~\cite{Goodfellow:NIPS14} and learns the domain-invariant features by fooling a discriminator which is to predict the domain of each~sample. Recently, contrastive/metric learning has also been proven effective for unsupervised domain adaptation~\cite{chen2021transferrable,Kang_2019_CVPR,Sharma2021InstanceLA,TMM_UDA_Wang_Contrastive,Pan2019TransferrablePN,pan2020exploring}.

\textbf{Explainability} is desired by AI to help the users understand the models' behaviors and diagnose the failures~\cite{arrieta2020explainable,core2006building,gunning2017explainable}. Deep models, which contain highly non-linear computation and inexplicable feature representations, are thus the least explainable. In \cite{craven1996extracting,thrun1995extracting}, researchers attempt to approximate the deep models with rules or linear classifiers, which are more interpretable to humans~\cite{swartout1981producing,swartout1993explanation}. Recently, a number of methods~\cite{Petsiuk2018rise,selvaraju2020grad,TMM_EXP_Wang,Elliott_2021_CVPR,Li_2021_ICCV,zeiler2014visualizing} improve the explainability of CNN on visual data via saliency maps, which highlight the pixels/regions that are crucial for model prediction. For the white-box models whose architectures and parameters are unconcealed, the gradient maps through back-propagation~\cite{selvaraju2020grad,zeiler2014visualizing,zhang2018top} and the learnable prototypes~\cite{alvarez-melis2018towards,chen2019this,li2017deep} are utilized to explain the models' predictions. For black-box models, the outputs of the perturbed images are investigated to convey rationale behind the predictions. The explainability of deep models is developed to further improve model predictions in \cite{bargal2018guided,cao2015look,selvaraju2019taking,zunino2021excitation}.

Despite having these progresses, the explainability of cross-domain recognition is not fully explored, which is the main theme of this paper. The related work of SFIT~\cite{Hou2021VisualizingAK} visualizes the adapted knowledge from the viewpoint of image style translation and Zunino \etal~\cite{zunino2021explainable} exploit the saliency-based explainability to improve the capability of domain generalization of deep classification models. Different from~\cite{Hou2021VisualizingAK,zunino2021explainable}, our work focuses on explaining the knowledge transfer for domain adaptation, by revealing the contribution of each source sample to the predictions of the samples in target domain.

\section{Interpretable Deep Classifier}
The main goal of Interpretable Deep Classifier (IDC) is to devise a novel classifier for cross-domain recognition whose decisions can be better understood.
To achieve this, IDC constructs a memory bank for each category, which stores the features of the discriminative source samples in the key and takes the corresponding representative scores for describing the category as the value. In the inference, IDC reads from the memory bank to measure the similarities between the target sample and the keys. The highly similar keys are then regarded as the evidence, whose similarities are utilized to weighted average representative scores as the prediction of the target sample and also indicate the contribution of source evidence to the target prediction.

\subsection{Explainable Decision-Making Process}
We begin by presenting the definition of an explainable decision-making process. In the traditional in-domain image recognition problem, the classifier is trained to predict the probability of the input image $x$ belonging to class $c$ as $P(y=c|x)$, where $y$ denotes the category of $x$. The classifier based on deep neural networks usually stack non-linear transformations to end-to-endly compute the probability. This process is opaque to the end users, and it is hard to explain the rationale behind the decisions. As such, the exploration of explainable deep models, whose predictions can be explained by intermediate evidence, has attracted extensive research attentions. Here, we decouple an explainable decision-making process into two steps: (i) given the input sample, identifying the evidence that is crucial for making decision, and (ii) inferring the probability of the input sample on each category w.r.t the evidence. The prediction can be calculated as:
\begin{equation}\label{eq:probability}
\small
\begin{aligned}
P(y=c|x) = \int\limits_{\mathcal{E}} \mathop{P(y=c|\mathcal{E})}_{inferring}\mathop{P(\mathcal{E}|x)}_{identifying}
\end{aligned}~~,
\end{equation}
where $\mathcal{E}$ denotes the evidence. The recent techniques on explaining deep models~\cite{Petsiuk2018rise,selvaraju2020grad} treat the most indicative image regions as the evidence $\mathcal{E}$ for inferring the category. In that case, the probability in Eq.(\ref{eq:probability}) can be re-written as an attention pooling across all candidate regions.

In our paper, we take a further step forward and study a more challenging scenario to explain unsupervised domain adaptation of image recognition. Suppose we have $N_s$ image-label pairs $\mathcal{X}_s=\{(x_s, y_s)\}$ in source domain and $N_t$ unlabeled images $\mathcal{X}_t=\{x_t\}$ in target domain. For unsupervised domain adaptation, the model is trained to transfer the knowledge from $\mathcal{X}_s$ to $\mathcal{X}_t$, and execute image recognition on target domain. To explain the rationale behind the model, we define the explainability in cross-domain recognition as the ability to assess the contribution of each source image to the predictions of target images. In other words, the evidence $\mathcal{E}$ in Eq.(\ref{eq:probability}) is defined as the most indicative source images for inferring the category of the target image. Formally, an explainable decision-making process for cross-domain image recognition could be formulated as
\begin{equation}\label{eq:cross-domain}
\small
\begin{aligned}
P(y_t=c|x_t) = \int\limits_{x_s \in \mathcal{X}_s} \mathop{P(y_t=c|\mathcal{E}=x_s)}_{inferring}\mathop{P(\mathcal{E}=x_s|x_t)}_{identifying}
\end{aligned}~~.
\end{equation}
By constructing the correlation between source images and the predictions of target images, this kind of process can nicely explain how the adaptive model transfers the knowledge across the source and target domains.

\subsection{Classification by Memory Matching in IDC}
\begin{figure}[t]
	\centering
	\includegraphics[width=0.99\linewidth]{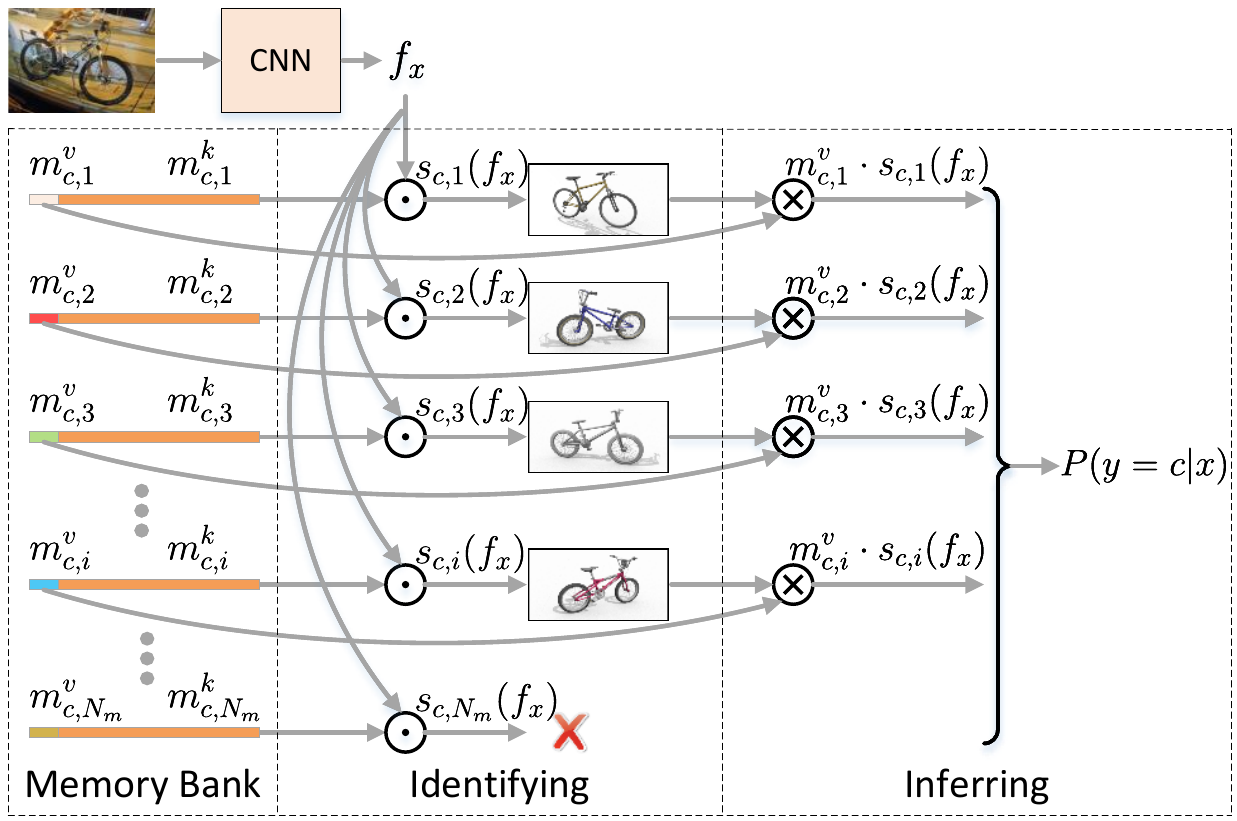}
	\caption{\small An overview of the memory matching based decision-making process in IDC. Given an input target image (e.g., bicycle), we perform identifying by calculating similarities $s_{c, i}(f_x)$ between query $f_x$ and keys $m_{c,i}^k$, and select the most similar source samples as the evidence. Inferring is then conducted based on the identified evidence and the probability $P(y=c|x)$ is obtained.}
	\label{fig.framework_memory}
\end{figure}
Inspired by the high explainability of $k$-nearest neighbors ($k$-NN) which categorizes the unlabeled image based on the nearest labeled images in the feature space, we derive the spirit of distance learning in $k$-NN and exploit a memory matching mechanism \cite{kaiser2017learning,miller2016key,pritzel2017neural,sukhbaatar2015end,weston2014memory,zhu2020inflated,Cai2018MemoryMN} in IDC.

\textbf{Memory bank.} The basic structure in our memory matching is memory bank, which stores the features of discriminative source images and the corresponding properties. Specifically, IDC consists of $C$ independent memory banks $\mathcal{M}=\{\mathcal{M}_c|c=1,2,\ldots,C\}$ and each corresponds to one specific category.
For each memory bank, there are $N_m$ memory slots to deposit the key-value pairs as $\mathcal{M}_c=\{(m_{c,i}^k, m_{c,i}^v)|i=1,\ldots,N_m\}$. The key $m_{c,i}^k\in\mathbb{R}^{D}$ denotes a $D$-dimensional feature of the source image from category $c$, and the value $m_{c,i}^v$ contains the learnable representative score of the source image for describing the category. We take the age of memory slot as an additional property, which represents the time passed away since the last reading of this slot. The age will be updated during training and is utilized to discard the low-frequency memory slot. Figure~\ref{fig.framework_memory} depicts an overview of memory matching in IDC.

\textbf{Identifying.} Given an input image $x$, we take the image representation extracted from a basic CNN as the query $f_x$ and perform the \textit{memory reading} operation in memory bank $\mathcal{M}_c$. \textit{Memory reading} calculates the similarity between the query $f_x$ and the features of source images stored in the memory bank, and identifies the most similar source samples as the evidence. Specifically, by measuring the query-key similarity via dot product, the similarity between query $f_x$ and $i$-th memory slot of memory bank $\mathcal{M}_c$ is given by
\begin{equation}\label{eq:similarity}
\small
\begin{aligned}
s_{c, i}(f_x)=(\frac{f_x\cdot m_{c,i}^k}{\|f_x\|\|m_{c,i}^k\|} + 1) / 2
\end{aligned}~~,
\end{equation}
where $\|\cdot\|$ denotes the L2-norm. The similarity is linearly normalized to $[0,1]$. We pick up $N_k$ memory slots whose keys are top-$N_k$ nearest neighbors among all the slots in one memory bank to the query and these memory slots are considered as the evidence upon the category prediction of the input image $x$.

\textbf{Inferring.} With the identified evidence (i.e., the most similar memory slots), the memory value $m_{c,i}^v$ is exploited as a learnable value to measure the confidence of each memory slot for inferring the category $c$. This value is treated as the representative score of the stored feature. The values of all memory slots are optimized by gradient descent during training. Hence, the probability of the input image $x$ belonging to class $c$ is computed by
\begin{equation}\label{eq:reasoning}
\small
\begin{aligned}
P(y=c|x) = \frac{1}{N_k}\sum_{i=1}^{N_k} m_{c,i}^v \cdot s_{c, i}(f_{x})
\end{aligned}~~,
\end{equation}
which summates the representative scores of the top-$N_k$ memory slots weighted by the corresponding similarity.

\textbf{Updating memory bank.}
During the training of IDC, we update the key-value pairs stored in the memory banks from two aspects, i.e., optimizing the memory values (representative scores) by gradient descent, and renewing key-value pairs in the memory banks by \textit{memory writing}.

\vspace{0.06in}\noindent (i) \emph{Optimizing memory values:} We first build a fully-connected layer to support the construction of loss function. Given a source training image $x$ with label $c$, the fully-connected layer outputs the probabilities $\mathbf{y}\in\mathbb{R}^{C}$ to pre-estimate the prediction on all categories w.r.t the query feature~$f_x$. Through this estimation, we can discover the most confusing negative category $\overline{c}$, which obtains the highest score other than the ground-truth label $c$ in $\mathbf{y}$. The optimization objective is to maximize the prediction of the ground truth category $c$, and meanwhile suppress the probability of the negative category $\overline{c}$. More specifically, the probability of the ground-truth category and negative category is obtained by performing \textit{memory reading} in memory bank $\mathcal{M}_{c}$ and $\mathcal{M}_{\overline{c}}$, respectively. Following~\cite{zhu2020inflated}, by measuring the loss via mean squared error (MSE), the loss of IDC is formulated as
\begin{equation}
   \small
   \label{eq:loss_idc}
   \begin{aligned}
      \mathcal{L}_{idc}=\text{MSE}(P(y=c|x), 1) + \text{MSE}(P(y=\overline{c}|x), 0)
   \end{aligned}~~.
\end{equation}
During optimization, the $m^k_{c,i}$ in memory bank are frozen to explicitly identify source images as evidence. IDC only back-propagates the differentiable loss $\mathcal{L}_{idc}$ to $m^v_{c,i}$ and regards $f_x$ and $m^k_{c,i}$ in Eq.(\ref{eq:reasoning}) as constants. Hence, the memory value can be optimized by gradient descent without a differentiable top-$N_k$ picking up operation in identifying. Such a desgin promises differentiable memory banks.

\vspace{0.06in}\noindent (ii) \emph{Renewing key-value pairs:} In addition to the gradient based optimization, IDC also renews the key-value pairs by \textit{memory writing} during training. Given the latest training image $x$ and its prediction $\mathbf{y}_c$ on the ground-truth category $c$ by the fully-connected layer, a new key-value pair is generated as $(f_x,\mathbf{y}_c)$.
\textit{Memory writing} operation puts the new pair into memory bank $\mathcal{M}_c$ by either storing the pair in an empty memory slot or overwriting the low-frequency key-value pair. In the case when there is no empty slot in memory bank $\mathcal{M}_c$, the key-value pair with the oldest age, i.e., the memory slot that has not be picked up for the longest time, is replaced by the new key-value pair. The age of each memory slot here is utilized to measure the reading frequency by training images. Nevertheless, if executing \emph{memory reading} only for source images, the age may be biased to the keys which are close to source images.
To tackle this issue, IDC implements \textit{memory reading} also on target images, and in view of the lack of labels, IDC reads every memory bank in answering each target image. Such operations refresh the ages of memory slots whose keys are highly similar to the features of target images as well.
In this way, the key-value pairs whose keys are close to either source or target images are more likely to be maintained, which are expected to be the most discriminative key-value~pairs.

\subsection{Joint Training with Adversarial Learning}
\begin{figure}[t]
    \centering
    \includegraphics[width=0.99\linewidth]{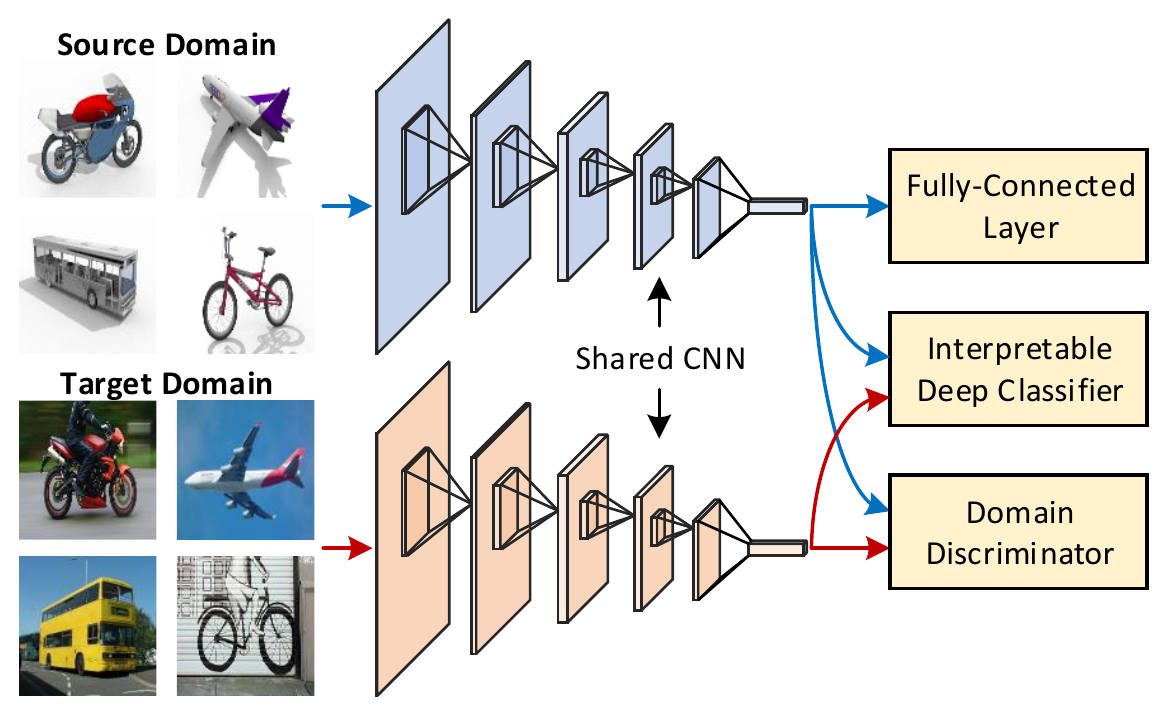}
    \vspace{-0.1in}
    \caption{\small An overall architecture of IDC training. We take a typical and widely-adopted adversarial learning framework as an example to show the joint training of IDC with UDA models.}
    \label{fig.framework_adv}
\end{figure}

\begin{algorithm}[!tb]\small
    \caption{\small The Training of Interpretable Deep Classifier}\label{alg:idc_train}
    \begin{algorithmic}[1]
        \STATE \textbf{Input:} \\
        ~~~~Source domain image-label pairs: $\mathcal{X}_s=\{(x_s, y_s)\}$. \\
        ~~~~Target domain images: $\mathcal{X}_t=\{x_t\}$. \\
        \STATE
        \textbf{Initialization:} \\
        ~~~~ImageNet pre-trained backbone network: $CNN$. \\
        ~~~~Random initialized domain discriminator: $D$. \\
        ~~~~Empty initialized Memory Banks: $\mathcal{M}$. \\
        \STATE
        \textbf{Output:} \\
        ~~~~Optimized parameters of $CNN$ and $\mathcal{M}$.
        \REPEAT
        \STATE
        Sample $(x_s, y_s)$ and $x_t$.
        \STATE
        Extract image representation: \\
        ~~~~$f_{x_s}=CNN(x_s)$, $f_{x_t}=CNN(x_t)$. \\
        \STATE
        Measure the probabilities $\mathbf{y}_{x_s}\in\mathbb{R}^{C}$ to pre-estimate the prediction of all categories w.r.t $f_{x_s}$.
        \STATE
        Identify the most confusing negative category $\overline{y_s}$, which achieves the highest score other than ground-truth $y_s$ in $\mathbf{y}_{x_s}$.
        \STATE
        Perform \textit{memory reading} with $f_{x_s}$ in $\mathcal{M}_{y_s}$ and $\mathcal{M}_{\overline{y_s}}$:\\
        ~~~~$P(y=y_s|x_s)=\textit{memory reading}(f_{x_s},\mathcal{M}_{y_s})$ \\
        ~~~~$P(y=\overline{y_s}|x_s)=\textit{memory reading}(f_{x_s},\mathcal{M}_{\overline{y_s}})$~.
        \STATE
        Perform \textit{memory reading} with $f_{x_t}$ in every memory bank: \\
        ~~~~~~~~$\textit{memory reading}(f_{x_t},\mathcal{M}_1)$ \\
        ~~~~~~~~$\textit{memory reading}(f_{x_t},\mathcal{M}_2)$ \\
        ~~~~~~~~$\cdots$ \\
        ~~~~~~~~$\textit{memory reading}(f_{x_t},\mathcal{M}_C)$~.
        \STATE
        Compute classification loss: \\
        ~~~~$\mathcal{L}_{fc}=L_{ce}(\mathbf{y}_{x_s}, y_s)$~.
        \STATE
        Compute adversarial loss: \\
        ~~~~$\mathcal{L}_{adv}(x_s, x_t)=-log(D(f_{x_t})) - log(1-D(f_{x_s}))$~.
        \STATE
        Compute IDC loss: \\
        ~~~~$\mathcal{L}_{idc}=\text{MSE}(P(y=y_s|x_s), 1)$\\
        ~~~~~~~~~~~~~~$ + \text{MSE}(P(y=\overline{y_s}|x_s), 0)$~.
        \STATE
        Perform \textit{memory writing} with $(f_{x_s},\mathbf{y}^{y_s}_{x_s})$ in $\mathcal{M}_{y_s}$.
        \STATE
        Jointly optimize $\mathcal{L}_{fc}$, $\mathcal{L}_{adv}$ and $\mathcal{L}_{idc}$.
        \STATE
        Back-propagate the gradients and update weights.
        \UNTIL{maximum iteration reached.}
    \end{algorithmic}
\end{algorithm}
  
\begin{algorithm}[!tb]\small
    \caption{\small The Inference of Interpretable Deep Classifier}\label{alg:idc_infer}
    \begin{algorithmic}[1]
        \STATE \textbf{Input:} \\
        ~~~~Image $x$.
        \STATE
        \textbf{Initialization:} \\
        ~~~~Optimized $CNN$, $\mathcal{M}$~.
        \STATE
        \textbf{Output:} \\
        ~~~~Prediction of image $x$~.
        \STATE
        \textbf{Procedure:} \\
        \STATE
        ~~~~Extract image representation: $f_x=CNN(x)$~.
        \STATE
        ~~~~Perform \textit{memory reading} with $f_x$ in every memory bank: \\
        ~~~~~~~~$P(y=1|x)=\textit{memory reading}(f_x,\mathcal{M}_1)$ \\
        ~~~~~~~~$P(y=2|x)=\textit{memory reading}(f_x,\mathcal{M}_2)$ \\
        ~~~~~~~~$\cdots$ \\
        ~~~~~~~~$P(y=C|x)=\textit{memory reading}(f_x,\mathcal{M}_C)$~.
        \STATE
        ~~~~Predict the category of $x$: \\
        ~~~~~~~~$\hat{y}=\mathop{\arg\max}_{1\le c \le C}P(y=c|x)$~.
    \end{algorithmic}
\end{algorithm}

As the core of IDC training, updating memory bank in IDC capitalizes on only source images labels, the prediction scores on source images w.r.t. all the categories, and source/target images features via CNN. 
Such information is available in most UDA works, e.g., MMD-based, adversarial learning based, and contrastive learning based methods, making IDC readily compatible with those models. For simplicity, here we take the widely-adopted adversarial learning framework as an example to integrate our IDC for unsupervised domain adaptation. 
\Cref{fig.framework_adv} illustrates the overall architecture. The principle behind this framework is equivalent to guiding the image recognition in both domains by making the representations from source and target images indistinguishable. Image representations are optimized via the fully-connected layer in a supervised manner by minimizing the classification loss on source images:
\begin{equation}
   \small
   \label{eq:loss_cls}
   \mathcal{L}_{fc}(\mathcal{X}_s)=\mathop{E}_{x_s\sim\mathcal{X}_s}L_{ce}(\mathbf{y}_{x_s}, y_s)~,
\end{equation}
where $E$ denotes the expectation over the set of images and $L_{ce}$ is the cross-entropy classification loss. $\mathbf{y}_{x_s}$ is the probability predicted by the fully-connected layer.
A domain discriminator $D$ is exploited to differentiate the features of source and target images. Meanwhile, the basic CNN is trained to maximally fool the discriminator through generating domain-invariant representations. The adversarial loss of the discriminator $D$ is
\begin{equation}
  \small
  \label{eq:loss_adv}
  \begin{aligned}
    \mathcal{L}_{adv}(\mathcal{X}_s, \mathcal{X}_t)\!=\!-\!\!\!\mathop{E}_{x_t\sim\mathcal{X}_t}[log(D(f_{x_t}))]\!-\!\!\!\mathop{E}_{x_s\sim\mathcal{X}_s}[log(1\!-\!D(f_{x_s}))]~.
  \end{aligned}
\end{equation}
The training of adversarial learning is a minmax game between the basic CNN and the discriminator, expecting a good equilibrium that the CNN can produce transferable representations after convergence. 
The overall objective of jointly optimizing IDC in adversarial learning framework can be written as
\begin{equation}
   \small
   \begin{aligned}
       \mathop{min}_{CNN,\mathcal{M}} \mathcal{L}_{fc}+\mathcal{L}_{idc}-\mathcal{L}_{adv},~~~~\mathop{min}_{D} \mathcal{L}_{adv}~,
   \end{aligned}
\end{equation}
where we empirically treat each loss equally and accumulate the three losses.
\cref{alg:idc_train} summarize the joint training of IDC with adversarial learning and \cref{alg:idc_infer} demonstrates the way to infer the class of an image by IDC.

\section{Experiments}
We empirically verify the merit of IDC for unsupervised domain adaptation on VisDA-2017~\cite{peng2017visda} and Office-Home~\cite{venkateswara2017deep} datasets. The first experiment compares our IDC with the classifier of the fully-connected layer in adversarial learning framework and examines the explanations or the evidence of source images for the decision-making process in IDC.
The second experiment regards IDC as an interpretable classifier to calibrate classification on target images with rejection. IDC capitalizes on the explanations as the cues and is able to choose not to make a prediction to avoid critical mistakes. The third experiment takes IDC as a prior interpreter to select the discriminative source images conditioning on the unlabeled target images, and executes the standard UDA on the selected source images plus unlabeled images from the target domain.

\subsection{Datasets}
\textbf{VisDA-2017}~\cite{peng2017visda} is to date the largest synthetic-to-real object classification dataset, containing more than 280K images in training, validation and test domains. There are 12 identical object categories across the three domains. The training domain includes 152K synthetic images generated by rendering 3D models of the object categories under various circumstances, such as different angles and lighting conditions. The validation domain consists of 55K objects cropped from real images in COCO~\cite{lin2014microsoft}. The test domain contains 72K images via cropping objects from video frames in YT-BB~\cite{real2017youtube}. The annotations of the images in test domain are not publicly available and here we conduct the experiments by considering the training domain as source and the validation domain as target. We report the averaged per-category classification accuracy as the metric. \textbf{Office-Home}~\cite{venkateswara2017deep} is a dataset created to assess domain adaptation for object recognition. The dataset includes over 15K web collected images from four different domains: 2,427 Artistic (Ar) images, 4,365 Clipart (Cl) images, 4,439 Product (Pr) images, and 4,357 Real-World (Rw) images. The images in each domain are from 65 object categories typically found in office and home scenarios. Following~\cite{venkateswara2017deep}, we evaluate 12 transfer directions of source to target selected from the four domains, and present the mean of accuracy on 12 directions as the performance on this dataset. On both datasets, we repeat each experiment at least eight times and report the mean and standard deviation.

\subsection{Implementation Details}
We implement our proposal in this paper on PyTorch~\cite{paszke2019pytorch} and exploit the ImageNet supervised pre-trained ResNet-50~\cite{he2016deep} structure as the backbone network. The number of memory banks in IDC and the number of memory slots in each bank depend on the datasets. For VisDA-2017, IDC has 12 memory banks and each bank contains $N_m=8,192$ memory slots. $N_k$ is set to 64 for \textit{memory reading} operation. For Office-Home with a relatively small number of images in each domain, IDC builds 65 memory banks and each bank includes $N_m=512$ slots. $N_k$ in this case is fixed to 4. 
Memory slots are empty initialized with negligible time consumption, and then filled/updated during training.
In the adversarial training, a gradient reversal layer (GRL)~\cite{ganin2015unsupervised} and the entropy conditioning~\cite{long2018conditional} are also applied for better domain adaptation.
IDC requires an extra 0.75/0.25 GiB device memory for VisDA/Office-Home and an additional 1.0/0.4$ms$ to process each image during training/inference on a V100~GPU.

For the network optimization in the adversarial domain adaptation framework, we employ the mini-batch stochastic gradient descent (SGD) with momentum 0.9 and weight decay $5\times 10^{-4}$. The learning rate and maximum training iterations are set as $3\times 10^{-4}$/$1\times 10^{-3}$ and 5$K$/4$K$ for transfer tasks on VisDA-2017/Office-Home datasets. The batch size is fixed to 72 and each mini-batch contains an equal number of images from source and target domain. To adjust the learnable representative scores in memory banks of IDC in training procedure, we use an Adam~\cite{kingma2015adam} optimizer with a small learning rate of $1\times10^{-5}$. For data augmentation, the input images are first resized to $256\times 256$ and then randomly cropped to $224\times 224$ from the resized images. The cropped version is also randomly horizontally flipped.

\begin{figure*}[!t]
    \centering\includegraphics[width=0.99\textwidth]{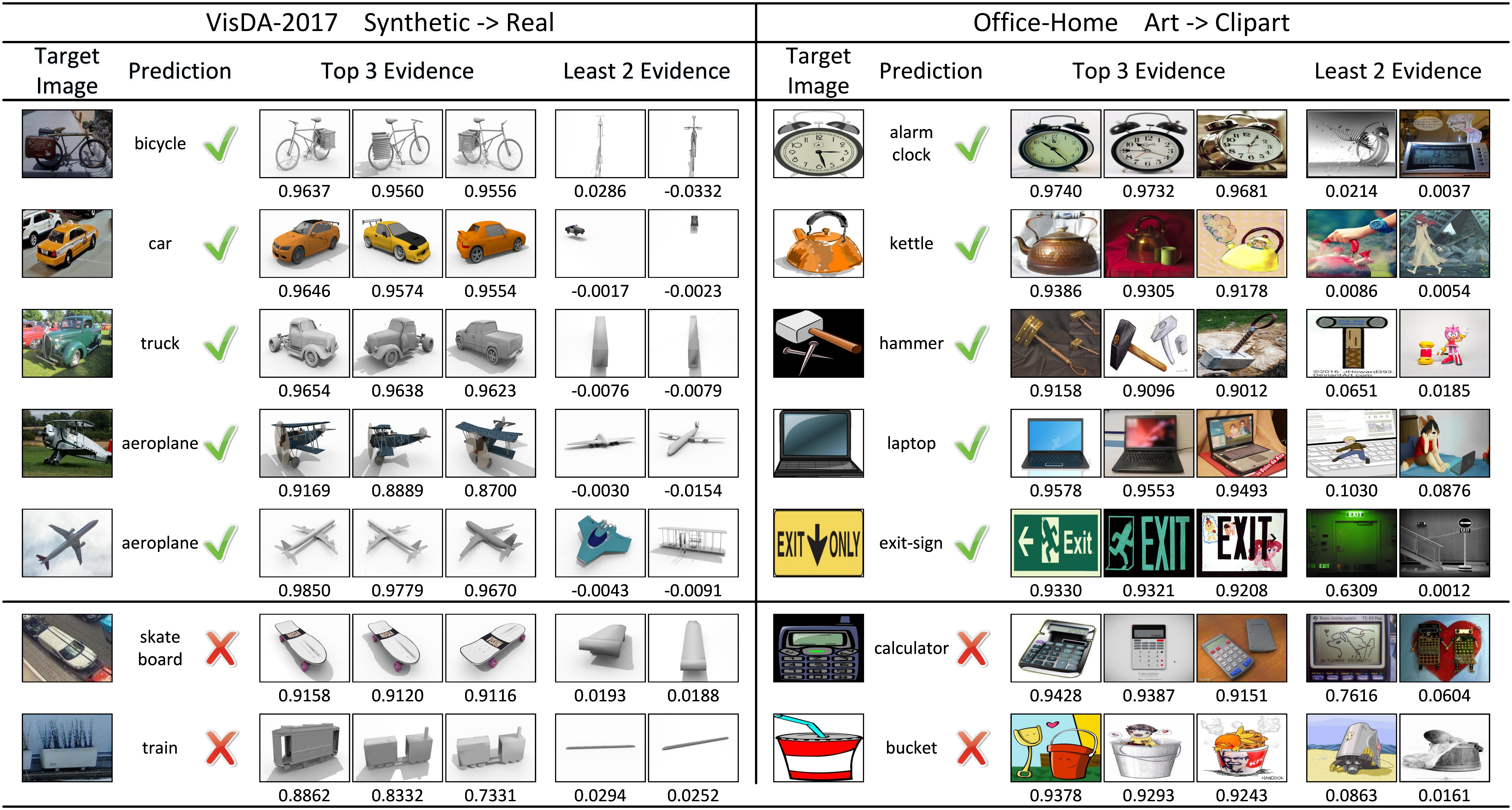}
    \vspace{-0.05in}
    \caption{\small Seven target images from each dataset with the predictions by IDC and the three/two source images that contribute the most/least to classify each target image.}
    \label{fig.exp_explain}
\end{figure*}

\subsection{Evaluations on IDC}
We first verify the effectiveness of IDC on UDA from two aspects: 1) does IDC maintain a high level of performance, and 2) how about the explanations or the evidence of the source images for IDC's predictions on target images.

\begin{table}[t]
  \centering
  \small
  \caption{\small Classification accuracy comparisons between fully-connected layer (FC) and IDC coupled with (a) adversarial learning and (b) transferable contrastive learning~\cite{chen2021transferrable}.}
  \label{tab:perf_idc}
  \vspace{-0.05in}
  \begin{subtable}[b]{0.9\linewidth}
    \centering
    \small
    \caption{\small Adversarial Learning (AL).}
    \vspace{-0.05in}
    \label{tab:perf_idc_adv}
    \begin{tabular}{ccc}
    \toprule
    Dataset & VisDA-2017      & Office-Home \\ \midrule
    FC      & 71.4$\pm$0.7 & 69.1$\pm$0.3 \\
    IDC     & 71.4$\pm$0.7 & 68.9$\pm$0.1 \\ \bottomrule
    \end{tabular}
    \vspace{0.1in}
  \end{subtable}
  \\
  \begin{subtable}[b]{0.9\linewidth}
    \centering
    \small
    \caption{\small Transferrable Contrastive Learning (TCL).}
    \vspace{-0.05in}
    \label{tab:perf_idc_tcl}
    \begin{tabular}{ccc}
    \toprule
    Dataset & VisDA-2017      & Office-Home \\ \midrule
    FC      & 89.1$\pm$0.5    & 73.0$\pm$0.4 \\
    IDC     & 89.3$\pm$0.3    & 73.1$\pm$0.3 \\ \bottomrule
    \end{tabular}
    \vspace{-0.1in}
  \end{subtable}
\end{table}

Without loss of generality, we consider the fully-connected (FC) layer coupled with the adversarial learning framework as a deep classifier, which is commonly used for cross-domain classification but the least explainable. We regard the classifier of FC layer as a good baseline and Table~\ref{tab:perf_idc}(a) compares the performances on both VisDA-2017 and Office-Home datasets. The results across the two datasets in consistently indicate that our explainable IDC achieves very comparable accuracy to that of FC. Moreover, IDC leads to a lower standard deviation of the performance on Office-Home and thus IDC is potentially more robust.
We also integrate IDC into the SOTA TCL model and IDC slightly outperforms FC (Table~\ref{tab:perf_idc}(b)), validating the effectiveness and compatibility of IDC.

\begin{figure*}[t]
    \centering\includegraphics[width=0.95\textwidth]{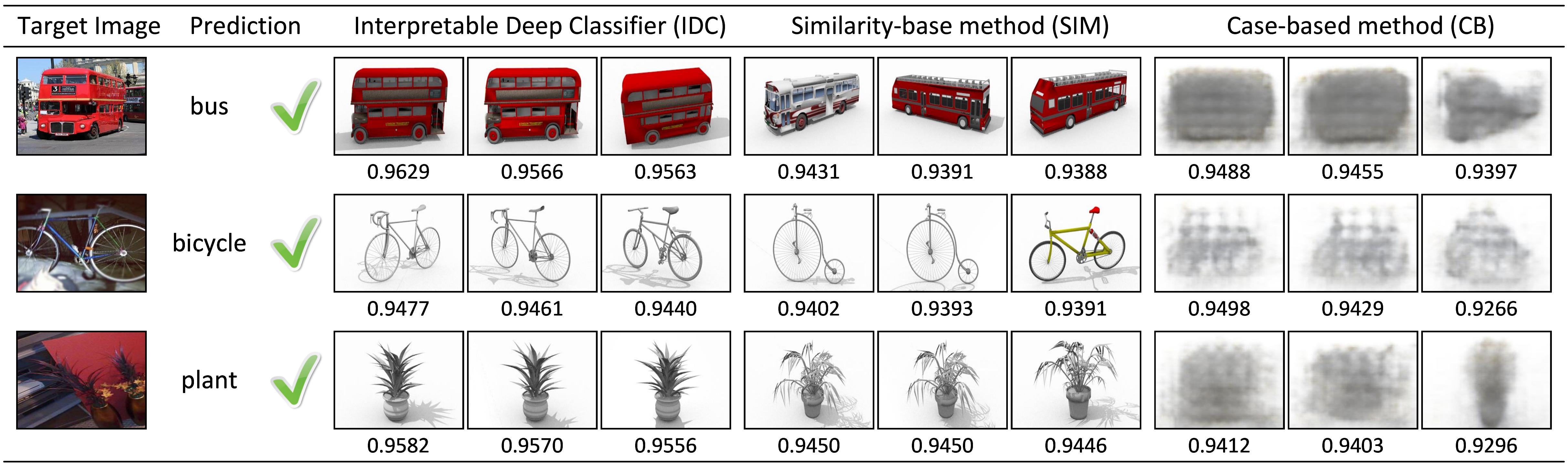}
    \vspace{-0.05in}
    \caption{\small Three target images from VisDA-2017 with predictions by IDC/SIM/CB and the three source images/prototypes that contribute the most to classify each target image.}
    \label{fig.exp_explain_visda_baselines}
    \vspace{-0.2in}
\end{figure*}

The explanations in IDC are defined as the evidence of source images upon which the classifier makes the predictions on target images. As such, we examine the quality of the explanations by looking into the evidence. Figure \ref{fig.exp_explain} showcases seven target images from each dataset with the predictions by IDC and also the three/two source images that contribute the most/least to classify the target image. The value associated with each source image is the product of the representative score of the source image and the similarity between the source and target images. The value could be negative, indicating that the corresponding source image is unable to describe the category well. Overall, IDC offers reasonable evidence of source images to explain the decisions on target images across the two datasets. The top-3 important source images are all visually similar to the target image in terms of either color, shape or style, no matter the prediction on target image is correct or not. Taking the target images from VisDA-2017 in the first two rows as examples, IDC nicely identifies the bicycles with a side basket or yellow cars in the source domain as the evidence to support the decision on target images with the specific property. In the meantime, the source images with the minimal contributions are quite distinct from target images and often capture the objects from an extreme view, e.g., a vertical view of bicycle or very small cars. Those source images are incapable of representing the object category and the representative scores become negative as the learning of IDC proceeds. IDC also shows good evidence on the other three target images from truck and aeroplane class. In particular, the two images in the fourth and fifth rows are both of aeroplane but they are different in visual appearance that the fourth one belongs to biplane and the fifth one is airliner. As indicated by our results, IDC successfully pinpoints the exact evidence of source images for recognizing the two target images. Furthermore, the lower part of two images illustrates the cases with wrong predictions by IDC. We speculate that this may be the result of the lack of objects in similar views with the target images in source domain and IDC has to rely on the visually similar samples from other categories as explanations. For instance, the white car in the vertical view or the flowerpot looks a bit like the skateboard or the train in source images. Even in these cases, the explanations of IDC are still seemingly reasonable. Similar results are observed on the Art$\rightarrow$Clipart transfer in Office-Home dataset, e.g., the correct evidence to recognize alarm clock, or visually similar evidence of calculator to misclassify the sample of telephone.

We also experimented with the similarity-based method (SIM) and the case-based method (CB)~\cite{li2017deep} to explain the cross-domain recognition on VisDA-2017.
The SIM measures the feature similarity between the target image and source images based on Eq.(\ref{eq:similarity}) and offers the nearest neighbors of the target images as evidence. The CB, which is an interpretable model originally designed for general image recognition, is remoulded for adversarial learning and gives the evidence by visualizing prototypes in the latent space. 
For the accuracy, IDC (71.4$\pm$0.7\%) performs better than SIM (70.3$\pm$0.5\%) and CB (69.0$\pm$1.0\%). Considering the explainability, \cref{fig.exp_explain_visda_baselines} showcases three target images classified by IDC/SIM/CB. For each target image, three source images that contribute the most to the recognition of each method are listed.
Benefiting from the learnable representative score in memory, the evidence offered by IDC is more reasonable. For example, IDC identifies the specific model of the double-deck bus in source domain which is not observed as the evidence in SIM and CB.

\subsection{IDC on Classifying Target Data with Rejection}
\begin{figure}[t]
\centering\includegraphics[width=0.99\linewidth]{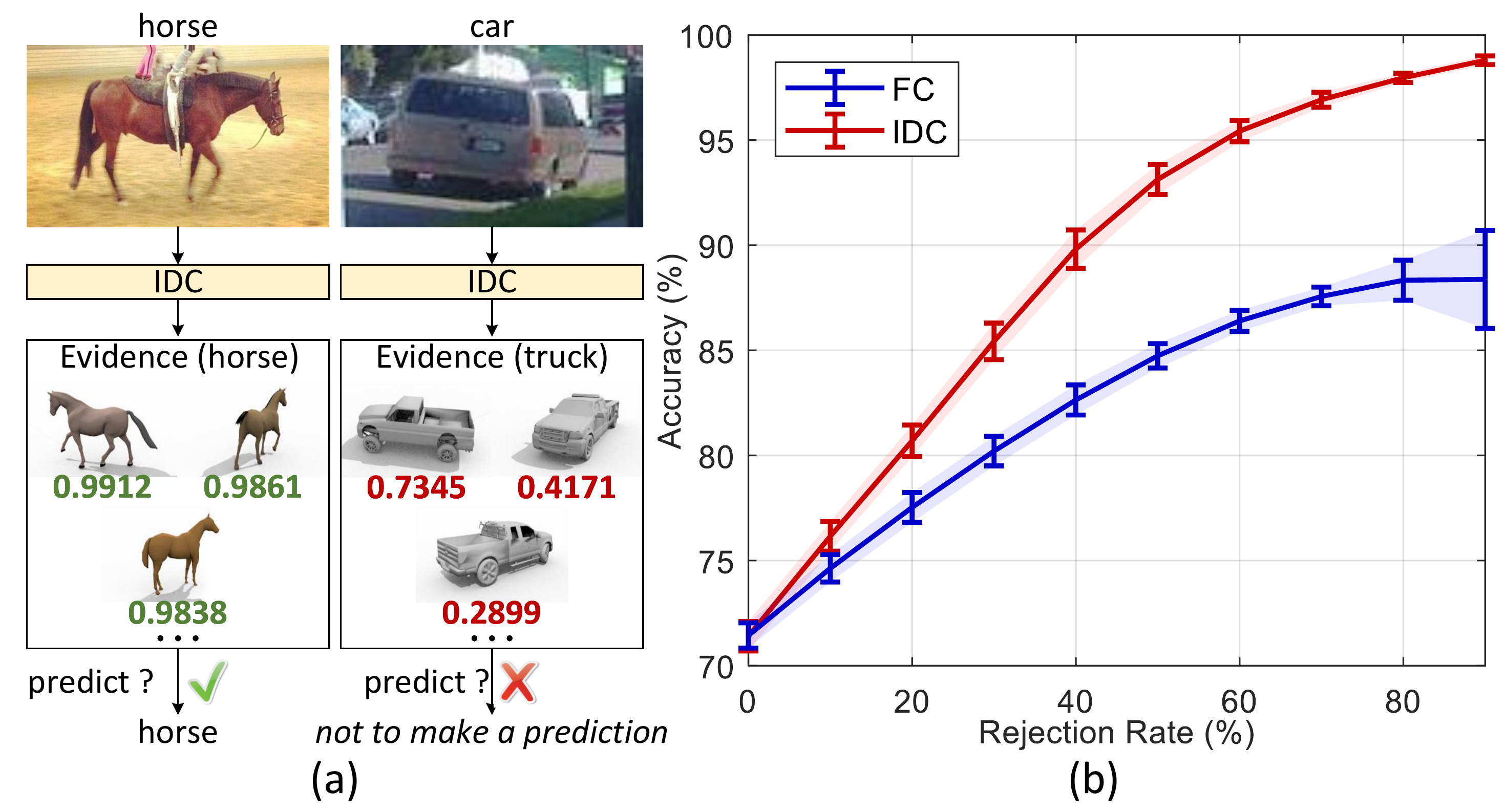}
\vspace{-0.1in}
\caption{\small (a) Two examples of classification with rejection by IDC. (b) Performance comparisons with different rejection rates.}
\label{fig.exp_rejection}
\end{figure}

\begin{table*}[!tb]
    \small
    \centering
    \caption{\small Comparisons on selecting different ratios of source images in the \textbf{VisDA-2017} dataset by various methods/strategies (all classifiers are trained in adversarial learning).}
    \label{tab:sample-selection-visda}
    \vspace{-0.08in}
    \begin{tabular}{l|cccccc|c}\toprule
    Ratio (\%) & 0.1      & 0.5      & 1        & 5        & 10       & 50       & 100        \\\midrule
    Random    & 64.6$\pm$1.3 & 71.3$\pm$0.5 & 71.4$\pm$1.2 & 71.6$\pm$1.1 & 71.4$\pm$0.9 & 71.5$\pm$0.9 &                               \\\cmidrule{1-7}
    IN-S      & 20.4$\pm$0.2 & 34.2$\pm$0.3 & 37.0$\pm$0.2 & 38.9$\pm$1.7 & 41.5$\pm$0.4 & 60.9$\pm$2.3 & \multirow{9}{*}{71.4$\pm$0.7} \\
    IN-P      & 57.7$\pm$2.3 & 66.5$\pm$0.8 & 68.4$\pm$0.1 & 72.3$\pm$1.4 & 71.9$\pm$1.8 & 72.2$\pm$0.5 &                               \\
    IN-M      & 62.9$\pm$1.6 & 69.8$\pm$0.3 & 72.7$\pm$2.0 & 73.1$\pm$0.5 & 73.9$\pm$0.2 & 73.5$\pm$0.6 &                               \\\cmidrule{1-7}
    ADV-S     & 28.1$\pm$4.4 & 33.0$\pm$0.1 & 36.2$\pm$1.8 & 39.0$\pm$1.1 & 39.9$\pm$1.1 & 53.9$\pm$1.1 &                               \\
    ADV-P     & 54.6$\pm$7.8 & 69.8$\pm$0.3 & 70.1$\pm$0.7 & 73.0$\pm$0.5 & 72.8$\pm$0.6 & 72.8$\pm$0.9 &                               \\
    ADV-M     & 63.8$\pm$2.2 & 73.2$\pm$1.0 & 73.9$\pm$1.2 & 75.6$\pm$0.5 & 74.8$\pm$0.6 & 74.0$\pm$1.1 &                               \\\cmidrule{1-7}
    IDC-S     & 59.5$\pm$3.6 & 60.8$\pm$3.1 & 74.7$\pm$0.7 & 73.8$\pm$0.4 & 73.0$\pm$1.2 & 66.8$\pm$2.0 &                               \\
    IDC-P     & 73.3$\pm$0.7 & 73.9$\pm$0.9 & 74.6$\pm$0.4 & 74.4$\pm$0.5 & 73.7$\pm$1.0 & 74.0$\pm$0.7 &                               \\
    IDC-M     & \textbf{76.0$\pm$1.5} & \textbf{76.0$\pm$0.6} & \textbf{76.7$\pm$0.7} & \textbf{76.1$\pm$0.8} & \textbf{74.9$\pm$0.5} & \textbf{75.3$\pm$1.0} \\\bottomrule
    \end{tabular}
    \vspace{-0.05in}
  \end{table*}
  
  \begin{table*}[!tb]
   \small
   \centering
  \caption{\small Comparisons on selecting different ratios of source images in the \textbf{Office-Home} dataset by various methods/strategies (all classifiers are trained in adversarial learning).}
   \label{tab:sample-selection-office}
   \vspace{-0.08in}
   \begin{tabular}{l|cccccc|c}\toprule
   Ratio (\%) & 10       & 20       & 30       & 40       & 60       & 80       & 100                                                     \\\midrule
   Random    & 53.4$\pm$0.5 & 61.4$\pm$0.1 & 64.3$\pm$0.2 & 66.0$\pm$0.1 & 67.6$\pm$0.2 & 68.3$\pm$0.1 &                                  \\\cmidrule{1-7}
   IN-S      & 31.9$\pm$0.2 & 44.7$\pm$0.1 & 53.2$\pm$0.1 & 58.3$\pm$0.1 & 64.3$\pm$0.1 & 66.7$\pm$0.1 &  \multirow{9}{*}{69.1$\pm$0.3}   \\
   IN-P      & 51.4$\pm$0.1 & 60.4$\pm$0.2 & 63.3$\pm$0.3 & 64.8$\pm$0.0 & 67.2$\pm$0.1 & 68.3$\pm$0.1 &                                  \\
   IN-M      & 55.7$\pm$0.1 & 63.0$\pm$0.0 & 66.2$\pm$0.1 & 67.2$\pm$0.1 & 68.9$\pm$0.3 & 69.4$\pm$0.2 &                                  \\\cmidrule{1-7}
   ADV-S     & 27.9$\pm$0.5 & 40.7$\pm$0.1 & 48.5$\pm$0.1 & 54.2$\pm$0.2 & 61.3$\pm$0.1 & 66.1$\pm$0.4 &                                  \\
   ADV-P     & 47.9$\pm$0.5 & 57.8$\pm$0.1 & 61.8$\pm$0.1 & 64.3$\pm$0.1 & 66.7$\pm$0.2 & 68.2$\pm$0.2 &                                  \\
   ADV-M     & 52.8$\pm$0.3 & 62.0$\pm$0.2 & 65.5$\pm$0.2 & 67.1$\pm$0.1 & 68.7$\pm$0.1 & 69.5$\pm$0.2 &                                  \\\cmidrule{1-7}
   IDC-S     & 54.4$\pm$0.1 & 64.4$\pm$0.2 & 67.0$\pm$0.0 & 68.0$\pm$0.3 & 68.2$\pm$0.2 & 68.3$\pm$0.1 &                                  \\
   IDC-P     & 51.2$\pm$0.2 & 59.8$\pm$0.0 & 63.6$\pm$0.1 & 65.7$\pm$0.2 & 67.5$\pm$0.1 & 68.4$\pm$0.0 &                                  \\
   IDC-M     & \textbf{56.9$\pm$0.2} & \textbf{65.2$\pm$0.1} & \textbf{67.7$\pm$0.1} & \textbf{68.7$\pm$0.0} & \textbf{69.8$\pm$0.2} & \textbf{69.6$\pm$0.1} &                            \\\bottomrule
   \end{tabular}
   \vspace{-0.15in}
  \end{table*}

In real-world deployment, misclassification can be costly. The task of classification with rejection is to prevent critical misclassification by providing an option not to make a prediction at the expense of the pre-defined rejection cost. Next, we assess the explanations in IDC through the task of classifying the target data with rejection on VisDA-2017. \cref{fig.exp_rejection}(a) illustrates two examples of classification with rejection. The score with each source image is the product of the representative score of this source image and the similarity between the source and target images. Such score indicates the confidence of classifying the target image from the viewpoint of this source image. For the target image of horse in the figure, all the evidence of source images by IDC shows high confidence scores and IDC then makes the correct prediction of horse. In contrast, due to large discrepancy between source and target domains in VisDA-2017, IDC wrongly provides the evidence of truck to the target image of car, but with low confidence scores. In this case, IDC chooses not to make the decision.

\cref{fig.exp_rejection}(b) quantitatively compares IDC with the fully-connected (FC) layer coupled with adversarial learning in terms of classification with rejection. Note that we calibrate FC with isotonic regression for fair comparisons. Moreover, we simply set the rate of ranking target images on their probability scores as the cost. There is an incentive to reject a target image with a relatively lower probability score. We rank the target images according to their probability scores from highest to lowest and reject the predictions on the images at different rates from the bottom of the ranking list. The run of IDC constantly yields better performances than FC across different rates and the red curve is always over the blue one. Especially, the improvement is more than 10\% if only predicting the target images with top 10\% probability scores, i.e., rejection rate of 90\%. The results again validate the explanations in IDC.

\subsection{IDC on Selecting Discriminative Source Data}
Finally, conditioning on the target data, we select the evidence of source samples by IDC to re-train the cross-domain models strating from ImageNet pre-trained weights for unsupervised adaptation and examine how the performance is affected. We compare the following four criteria for sample selection. \textbf{Random} selects the source images randomly irrespective of the evidence for target domain and usually acts as a basic method for data sampling. \textbf{IN} employs the ImageNet supervised pre-trained ResNet-50 to extract image features and measures the importance of each source image through averaging the similarities between the image and every target image in feature space. \textbf{ADV} shares the same process to estimate the importance of source images with \textbf{IN} but instead uses the ResNet-50 fine-tuned in adversarial learning framework for domain adaptation on corresponding dataset. \textbf{IDC} takes our IDC as a prior interpreter. Given a target image, \textbf{IDC} multiplies the representative score of one source image by the similarity between the source image and the target one to compute the importance of the source image in regard to that target image. The importance of the source image with respect to the target domain is then calculated by averaging the importance scores over all target images. In other words, all target images are regarded as the references for measuring the importance of each source image.

\begin{figure}[t]
\centering
\includegraphics[width=0.99\linewidth]{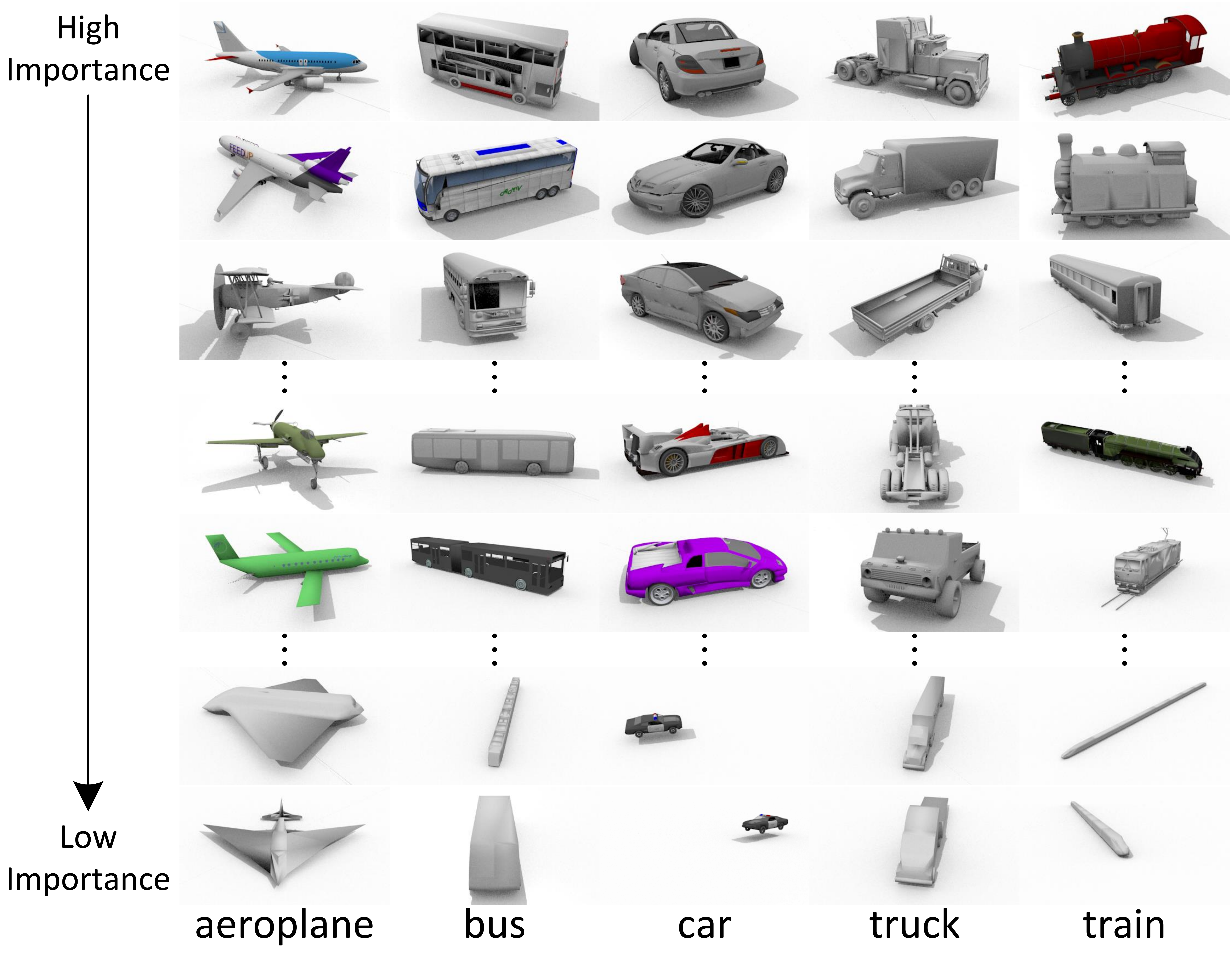}
\caption{\small The image lists of five categories in VisDA-2017 in the order of importance measured by IDC.}
\label{fig.exp_visualize_selection}
\vspace{-0.20in}
\end{figure}
      
\begin{table*}
\centering
\small
\caption{\small Performances of UDA models on the selections by IDC-M on VisDA-2017.}
\label{tab:selection-methods}
\begin{tabular}{c|cccccc|c}\toprule
\%       & 0.1      & 0.5      & 1        & 5        & 10       & 50       & 100      \\\midrule
AL   & 76.0$\pm$1.5          & 76.0$\pm$0.6          & \textbf{76.7$\pm$0.7} & {\ul 76.1$\pm$0.8} & 74.9$\pm$0.5 & 75.3$\pm$1.0 & 71.4$\pm$0.7 \\
DAN~\cite{long2015learning}  & 62.7$\pm$1.5          & 65.0$\pm$1.2          & \textbf{65.7$\pm$1.0} & {\ul 65.1$\pm$1.1} & 64.1$\pm$0.6 & 63.1$\pm$0.8 & 61.7$\pm$0.9 \\
DANN~\cite{ganin2016domain} & \textbf{77.6$\pm$1.4} & {\ul 76.9$\pm$1.2}    & 76.2$\pm$1.0          & 74.2$\pm$0.9       & 72.9$\pm$0.9 & 73.0$\pm$0.5 & 72.4$\pm$0.6 \\
JAN~\cite{long2017deep}  & {\ul 70.9$\pm$1.1}    & \textbf{71.0$\pm$0.7} & 70.2$\pm$0.5          & 68.5$\pm$0.5       & 67.4$\pm$0.6 & 67.3$\pm$0.5 & 65.4$\pm$0.5 \\
CDAN~\cite{long2018conditional} & 75.4$\pm$2.4          & \textbf{76.6$\pm$1.3} & {\ul 75.5$\pm$0.8}    & 74.7$\pm$0.7       & 74.1$\pm$0.4 & 74.3$\pm$0.7 & 74.2$\pm$0.7 \\
MCD~\cite{saito2018maximum}  & 70.3$\pm$1.3          & \textbf{72.8$\pm$0.9} & {\ul 72.6$\pm$0.9}    & 71.3$\pm$0.7       & 71.2$\pm$0.6 & 70.9$\pm$0.6 & 69.0$\pm$0.9 \\
MDD~\cite{zhang2019bridging}  & 76.1$\pm$2.9          & {\ul 78.2$\pm$3.0}    & \textbf{78.3$\pm$2.2} & 77.5$\pm$1.5       & 75.8$\pm$0.9 & 75.9$\pm$1.1 & 74.0$\pm$2.1 \\
CAN~\cite{Kang_2019_CVPR}  & \textbf{87.6$\pm$0.1} & {\ul 87.5$\pm$0.2} & 87.4$\pm$0.1 & 87.3$\pm$0.2 & 87.3$\pm$0.3 & 87.1$\pm$0.3 & 87.0$\pm$0.2 \\
TCL~\cite{chen2021transferrable}  & 89.4$\pm$0.2 & {\ul 89.5$\pm$0.2} & \textbf{89.7$\pm$0.1} & 89.4$\pm$0.1 & 89.4$\pm$0.3 & 89.2$\pm$0.3 & 89.1$\pm$0.5 \\ \bottomrule
\end{tabular}
\vspace{-0.1in}
\end{table*}

Based on the importance measured in \textbf{IN}, \textbf{ADV} and \textbf{IDC}, we also compare three strategies for selection. \textbf{-S} chooses the source images \textbf{s}olely depending on their importance from high to low scores until the number of selections is met. \textbf{-P} additionally preserves the original \textbf{p}roportion of samples across different categories in source domain to select source images. \textbf{-M} is a \textbf{m}ixture strategy to balance the samples across categories and exhaustively select samples by \textbf{-S}. \textbf{-M} splits the quotas into two parts. One part (e.g., 90\% in our experiments) is evenly distributed to choose the most important source images from each category. The other part (i.e., 10\%) is to collect the important ones from the remaining source images irrespective of the~category.

We re-train the cross-domain classifiers on the source selections and unlabeled target images in the adversarial learning (AL) framework. Table \ref{tab:sample-selection-visda} and \ref{tab:sample-selection-office} summarizes the performances on VisDA-2017 and Office-Home when using different quotas for selections, i.e., the ratio of source selections to all source images. Specifically, 
on the synthetic$\rightarrow$real tranfer in VisDA-2017,
unsupervised domain adaptation with 10\% source selections by Random obtains comparable accuracy to that with all source images. This is actually not surprising because 10\% source data in VisDA-2017 still includes more than 1,000 source images of each category, which may be sufficient for cross-domain learning in AL framework.

\textbf{-M} strategy takes the advantages of both the balance between categories and the importance holistically in source domain, leading to consistent improvements over \textbf{-S} and \textbf{-P} across three measures of importance. \textbf{IDC} benefits from the learning of representativeness of source images and constantly exhibits better performance than \textbf{IN} and \textbf{ADV}. In particular, utilizing only 0.1\% source data selected by \textbf{IDC} with \textbf{-M} strategy achieves superior accuracy than that uses all source data, which is very impressive. The results somewhat reveal the large gap between the two domains and directly enforcing the distribution of all source data to match the target may result in negative transfer in this case. In contrast, \textbf{IDC} selects the important source images with respect to the target domain and thus alleviates the gap in between. \Cref{fig.exp_visualize_selection} shows the image lists of five categories in the order of importance measured by \textbf{IDC}. Similar performance trends are observed on domain adaptation in Office-Home (Table \ref{tab:sample-selection-office}). \textbf{IDC-M} leads to performance boost against other selections across different ratios. With the selections of 60\%/80\% source images, \textbf{IDC} outperforms domain adaptation with all source data in the adversarial learning manner.

Table \ref{tab:selection-methods} further details the performances when training different UDA models on the selections by \textbf{IDC} with \textbf{-M} on VisDA-2017. As indicated by the results, all UDA models learnt on the selections of only 0.1\% source data outperform those trained on full source data. Specifically, training the state-of-the-art model TCL on the 0.1\% source data selected by IDC boosts up the accuracy from 89.0$\pm$0.6\% to 89.4$\pm$0.2\%. Increasing the selections to 0.5\% and 1\% will attain larger performance gains for most methods. In other words, we can simply conduct the process of ``train IDC$\rightarrow$select data$\rightarrow$train UDA models'', and obtain an accuracy boost for cross-domain recognition. The results clearly verify the selections by IDC, thereby demonstrating the effective explainability of IDC.

\section{Conclusions}
We have presented Interpretable Deep Classifier (IDC) which explains the evidence behind the cross-domain recognition. Particularly, we study the problem by delving into the high explainability of $k$-NN and derive from the spirit of distance learning. To materialize our idea, we build a memory bank for each category. The memory slot takes the image features as the key and the properties, e.g., the learnable representative scores of the features, as the value. IDC reads from the memory bank to identify the evidence of source images for recognizing the target image, based on the representative scores of the evidence and the similarity between evidence and the target image. Experiments conducted on both VisDA-2017 and Office-Home datasets validate our proposal and IDC learns effective explanations while maintaining the performance. Capitalizing on the explanations as the cues, IDC also demonstrates a high capability of preventing critical misclassification. More remarkably, performing unsupervised domain adaptation on 0.1\% source data selected by IDC achieves the superior accuracy than that uses all source data on the VisDA-2017 dataset.

{\small
\bibliographystyle{ieee_fullname}
\bibliography{egbib}
}

\end{document}